\def\year{2019}\relax
\documentclass[letterpaper]{article} 
\usepackage{aaai19}  
\usepackage{times}  
\usepackage{helvet}  
\usepackage{courier}  
\usepackage{url}  
\usepackage{graphicx}  
\begin{document}
%
\title{Non-Local Context Encoder: Robust Biomedical Image Segmentation against Adversarial Attacks}
\author{Xiang He$^1$, Sibei Yang$^2$, Guanbin Li$^1$\thanks{Xiang He and Sibei Yang contributed equally to this work. Corresponding author is Guanbin Li~(Email: liguanbin@mail.sysu.edu.cn). This work was partially supported  by the National Natural Science Foundation of China under Grant No.61702565 and the Fundamental Research Funds for the Central Universities under Grant No.18lgpy63.}, Haofeng Li$^2$, Huiyou Chang$^1$, Yizhou Yu$^3$\\
$^1$School of Data and Computer Science, Sun Yat-sen University, China\\
$^2$The University of Hong Kong, Hong Kong\qquad$^3$Deepwise AI Lab, China
}
\maketitle
\begin{abstract}
Recent progress in biomedical image segmentation based on deep convolutional neural networks (CNNs) has drawn much attention. However, its vulnerability towards adversarial samples cannot be overlooked. This paper is the first one that discovers that all the CNN-based state-of-the-art biomedical image segmentation models are sensitive to adversarial perturbations. This limits the deployment of these methods in safety-critical biomedical fields. In this paper, we discover that global spatial dependencies and global contextual information in a biomedical image can be exploited to defend against adversarial attacks. To this end, non-local context encoder (NLCE) is proposed to model short- and long-range spatial dependencies and encode global contexts for strengthening feature activations by channel-wise attention. The NLCE modules enhance the robustness and accuracy of the non-local context encoding network (NLCEN), which learns robust enhanced pyramid feature representations with NLCE modules, and then integrates the information across different levels. Experiments on both lung and skin lesion segmentation datasets have demonstrated that NLCEN outperforms any other state-of-the-art biomedical image segmentation methods against adversarial attacks. In addition, NLCE modules can be applied to improve the robustness of other CNN-based biomedical image segmentation methods.
\end{abstract}

\section{Introduction}
\noindent Biomedical image analysis catches people's eyes due to its popular application in computer-aided diagnosis and medical plan recommendation. Biomedical image segmentation is fundamental in biomedical image analysis, which performs pixel-level annotation for regions of interest (e.g. organs, substructures, and lesions) on biomedical images (e.g. X-ray, Magnetic Resonance Imaging, Computerized Tomography). However, it is challenging to obtain accurate segmentation because of the large shape and size variations of regions of interest, and the diversity of images produced by different biomedical imaging equipments \cite{hwang2017accurate,sarker2018slsdeep}. State-of-the-art biomedical image segmentation methods are based on fully convolutional networks (FCN) \cite{long2015fully}, which is a type of deep convolutional neural networks (CNNs) designed for semantic segmentation in computer vision. The accuracy of CNN-based biomedical image segmentation has been beyond that of traditional ones~\cite{litjens2017survey,sarker2018slsdeep,hwang2017accurate,ronneberger2015u,novikov2018fully,yuan2017automatic}. In addition to the accuracy of biomedical image segmentation, its stability and robustness are also essential for the fault-free clinical practice.

\begin{figure}[t]
  \centering
  \includegraphics[scale=0.35]{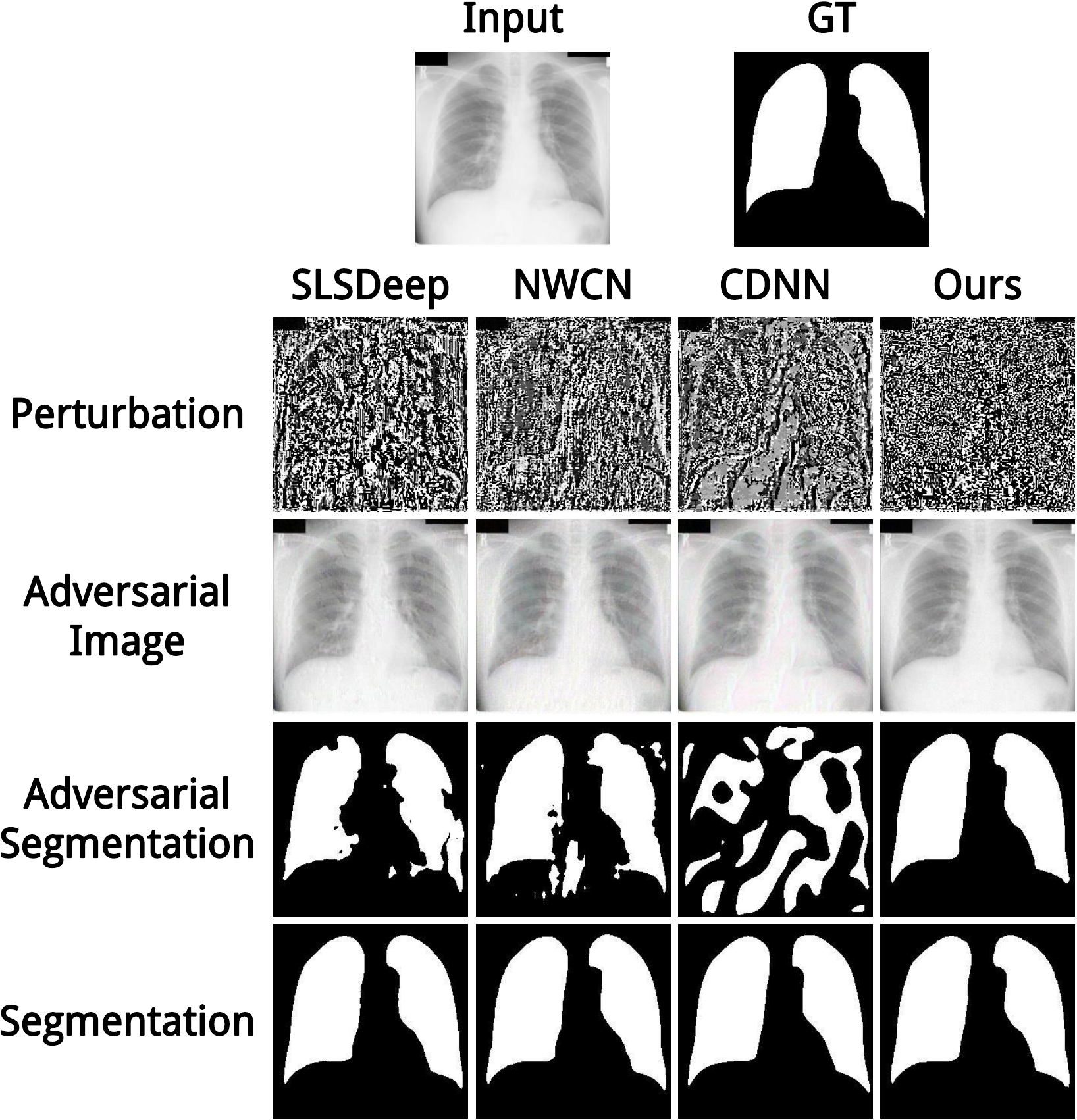}
  \caption{Sample adversarial attacks on SLSDeep \cite{sarker2018slsdeep}, NWCN \cite{hwang2017accurate}, CDNN \cite{yuan2017automatic} and our NLCEN. The input chest radiograph and its ground-truth segmentation are shown in the first row. Adversarial perturbations and images generated for models by the Iterative FGSM attack method \cite{kurakin2016adversarial} with adversarial intensity set to 16 are shown in the second and third rows respectively. Segmentation results on the adversarial images and the input image are shown in the fourth and fifth rows respectively.\vspace{-3mm}}
  \label{fig:I}
\end{figure}


Although CNN-based methods excel in solving many visual recognition tasks~\cite{lecun2015deep,ren2015faster,li2017instance,li2018contrast,li2018cross}, the vulnerability of CNNs to adversarial attacks cannot be overlooked~\cite{szegedy2013intriguing}. Adversarial samples are legitimate samples with human-imperceptible perturbations, which attempt to fool a trained model to make incorrect predictions with high confidence~\cite{szegedy2013intriguing}.
Such human-imperceptible perturbations, that CNNs are very sensitive to, are called adversarial noise. By exploiting the gradient-based error back-propagation mechanism for CNN training, adversarial attacks generate adversarial noise in an input image by back-propagating the error gradient induced by an intended incorrect prediction through a trained CNN model.

Recent work shows that complex semantic segmentation models, which are trained with an independent cross-entropy loss at each pixel on an image, are threatened by adversarial attacks~\cite{xie2017adversarial,arnab2017robustness}. Although biomedical image segmentation models share a similar deep learning framework with semantic segmentation models, adversarial attacks targeted at them have not been well explored. Since biomedical image segmentation does not have sufficient high-quality training samples, trained models can easily experience overfitting and exhibit a weak generalization capability, which make them more sensitive to noise. This property makes the models more vulnerable when facing adversarial attacks, and challenges their use in safety-critical biomedical fields. An example of adversarial attacks on lung segmentation is shown in Figure~\ref{fig:I}. 

The common defense strategy against adversarial attacks is adversarial training, which injects adversarial samples into training data to improve the robustness of trained models. \citeauthor{tramer2017ensemble} \citeyear{tramer2017ensemble} show that if the adversarial samples are taken as augmented data, the consequence of adversarial attacks can be alleviated. However, this defense strategy is limited because the adversarial samples are obtained from specific models and their corresponding adversarial attack methods. Therefore, instead of a limited training strategy, we wish to design a generic module, which can be easily integrated into CNN-based biomedical image segmentation networks to improve their robustness.

The robustness of biomedical image segmentation can be improved effectively by global spatial dependencies and global contextual information. Therefore, we propose to model them with a module called non-local context encoder (NLCE). In order to better introduce the effectiveness of global spatial dependencies and global contexts, we use a single pixel as an example, and the situation of a single pixel can be easily extended to the entire image because segmentation models are trained with independent loss at every pixel. First, global spatial dependencies are very important in defending against adversarial attacks. Given a pixel, capturing its global spatial dependencies means finding all highly related pixels within the entire image, and the prediction at this pixel is affected by all those pixels. There are two perspectives to understand the effectiveness of global dependencies. One is that if an incorrect label was given to a pixel, the incorrect loss at the pixel would be passed to all other related pixels by back-propagation, which increases the intensity of perturbation, and makes the adversarial sample significantly different from the original image. The other is that the noise at a pixel can be gradually weakened by the fusion with its highly related pixels in the process of forward-propagation. Second, global contextual information has a positive effect in defending against adversarial attacks because the configuration of the human body is relatively stable. For example, in lung image segmentation, the left and right lungs provide geometric contextual information by learning their geometric relationship with respect to each other. Because of the association between the left and right lungs, the right lung needs to receive the same perturbation-based attacks when the left lung is attacked. Therefore, the intensity of the required perturbation is increased.
Unfortunately, on one hand, CNNs have difficulty in capturing global dependencies because convolution operations only capture short-range dependencies by processing one local neighborhood at a time. Although stacked convolution operations are capable of capturing long-range dependencies by enlarging receptive fields \cite{fukushima1980neocognitron:,lecun1989backpropagation}, they increase the difficulty of optimization and may face the problem of gradient vanishing. On the other hand, current biomedical image segmentation methods do not make full use of global contextual information.

Inspired by the above analysis, in this paper, we propose a robust non-local context encoder module for biomedical image segmentation. The NLCE module captures the global spatial dependencies within a feature map by obtaining the response at a position of the feature map as a weighted sum of the features at all positions, and strengthens the features with channel-wise attention computed from the encoded global contextual information. In principle, the proposed robust NLCE module can also be applied to all CNN-based biomedical image segmentation methods and is able to improve the robustness of these models against adversarial attacks.

Moreover, we design and implement a medical image segmentation framework, named non-local context encoding network (NLCEN), which consists of two phases, the global phase and the refinement phase. Our global network is based on the feature pyramid network (FPN)~\cite{lin2017feature} and our NLCE modules. It learns global feature representations at different levels. The refinement network fuses features at different levels to obtain sharp boundaries. We conduct experiments on two common benchmark biomedical image segmentation datasets, the JSRT dataset for lung segmentation~\cite{shiraishi2000development} and the ISBI 2016 dataset~\cite{gutman2016skin} for skin lesion segmentation. Experimental results show that our NLCEN with NLCE modules has both high segmentation accuracy and robustness against adversarial attacks, and the NLCE modules practically help improve the segmentation accuracy of other biomedical image segmentation methods when they face adversarial attacks.

In summary, this paper has the following contributions:
\begin{itemize}
\item This is the first paper, to the best of our knowledge, attempts to improve the robustness of biomedical image segmentation methods by adding a robust module to the network. It proposes to exploit global spatial dependencies and global contexts to effectively improve the robustness of biomedical image segmentation methods.
\item It proposes non-local context encoder (NLCE), which is a robust biomedical image segmentation module against adversarial attacks. The NLCE module is able to capture distance-independent dependencies and global contextual information. And it can be easily applied to other CNN-based image segmentation methods.
\item It introduces non-local context encoding network (NLCEN), which achieves high segmentation accuracy and is robust on adversarial samples with different levels of adversarial perturbations.
\end{itemize}

\section{Related Work}
\subsection{Biomedical Image Segmentation}
The state-of-the-art biomedical image segmentation methods have similar frameworks to CNNs-based semantic segmentation models, but with fewer convolutional blocks and fewer network parameters to avoid overfitting. The U-net network is the most well-known segmentation method for biomedical image segmentation, and it is based on FCNs, but its upsampling phase and the downsampling phase use the same number of convolution operations in each level and the skip connection is used to connect the downsampling layer to the upsampling layer \cite{ronneberger2015u}. InvertedNet is an improved version of U-net that has fewer parameters to reduce overfitting, and for more accurate localization, it adopts delayed subsampling and learns higher resolution features \cite{novikov2018fully}. In order to use contextual information while maintaining resolution, NWCN adopts an atrous convolution-based model and utilizes a multi-stage training strategy to refine the preliminary segmentation results \cite{hwang2017accurate}. CDNN is also based on FCNs and it designs a loss function based on Jaccard distance \cite{yuan2017automatic}. SLSDeep, consisting of skip-connections, dilated residual and pyramid pooling, is an efficient skin lesion segmentation model from dermoscopic images. Its loss function, including negative log likelihood and end point error loss, is designed to obtain sharp boundary \cite{sarker2018slsdeep}.

\subsection{Adversarial Attacks}
Since the adversarial attacks to deep neural networks have been proposed by \citeauthor{szegedy2013intriguing}, they have received extensive attention. They are designed to generate the adversarial samples to fool a trained model to make incorrect predictions with high confidence. The adversarial perturbation is estimated by solving penalized optimization problem by using L-BFGS optimization method \cite{szegedy2013intriguing}. \citeauthor{goodfellow2015explaining} believe that the main reason why neural networks are vulnerable to adversarial attack is their linear behavior in high-dimensional space and propose the single-step fast gradient sign method (FGSM) to generate adversarial samples directly and efficiently. The single-step targeted attack is a modified version of FGSM, which aims at reducing the loss function of target category instead of the increasing the loss function of the original category \cite{kurakin2016adversarial}. In addition, the proposed basic iterative method can increase the success rate of attacks. The adversarial samples generated by iterative methods are less transferable than those generated by single-step attacks \cite{kurakin2016adversarial,arnab2017robustness}. \citeauthor{xie2017adversarial} are the first to explore adversarial attacks on image segmentation and detection on large datasets and propose the density adversary generation to generate effective adversarial samples by considering all the targets simultaneously. \citeauthor{arnab2017robustness} present the first rigorous evaluation on the robustness of the state-of-the-art semantic segmentation models to single-step adversarial attacks and iterative adversarial attacks.

\subsection{Global Modeling}
The global information modeling of images is an important part of the visual recognition field, and global information is utilized in many visual recognition tasks, e.g. scene segmentation \cite{li2016lstm},  saliency detection \cite{li2016visual,li2017instance} and semantic segmentation \cite{zhang2018context}. Getting global image information for CNN-based models is challenging, and it needs to consider both local dependencies and long-range dependencies. Stacked convolutional blocks can only capture local information due to restricted receptive fields. LSTM-CF treats spatial feature map obtained by CNNs as horizontal and vertical sequences respectively. It adopts multiple bi-directional long short term memory networks (LSTMs) in vertical direction to capture vertical short and long-range context, then the context is fused to get global spatial information by applying another bi-directional LSTMs in horizontal \cite{li2016lstm}. However, recurrent operations, like LSTMs, are still progress a local neighbor at a time, and the connection between two distant points must pass through the intermediate points. To capture the long-distance dependency, \cite{wang2018non} proposed a fast and direct method, which considers the features at all the positions to capture the dependencies at a position in a low-level feature map. \citeauthor{zhang2018context} takes the entire dataset into account and learns a set of global inherent representative of features to capture the global context for images. The global information for a feature map is obtained by encoding the relationships between its all features and the representative features.

\begin{figure*}[h]
  \begin{center}
  \includegraphics[width=0.9\linewidth]{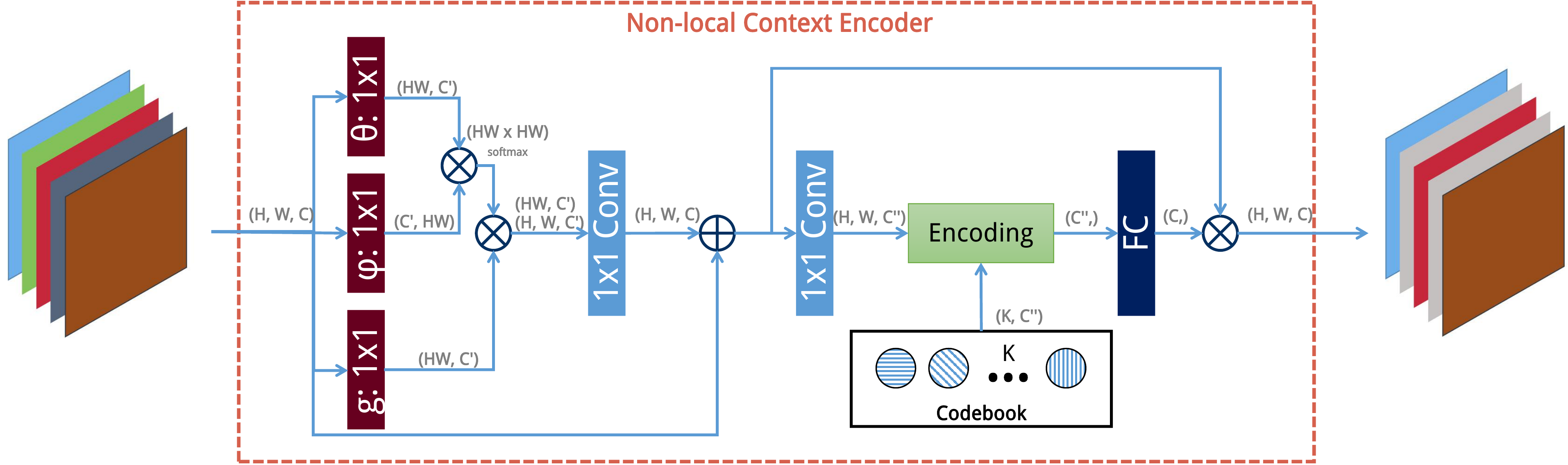} 
  \end{center}
  \caption{The architecture of our proposed non-local context encoder (NLCE). Our NLCE module first enhances and denoises the feature map by modeling global spatial dependencies and then applies channel-wise feature map attention by using encoded global context computed from a learned codebook.}
  \label{fig:nlce}
\end{figure*}

\begin{figure*}[h]
  \begin{center}
  \includegraphics[width=0.9\linewidth]{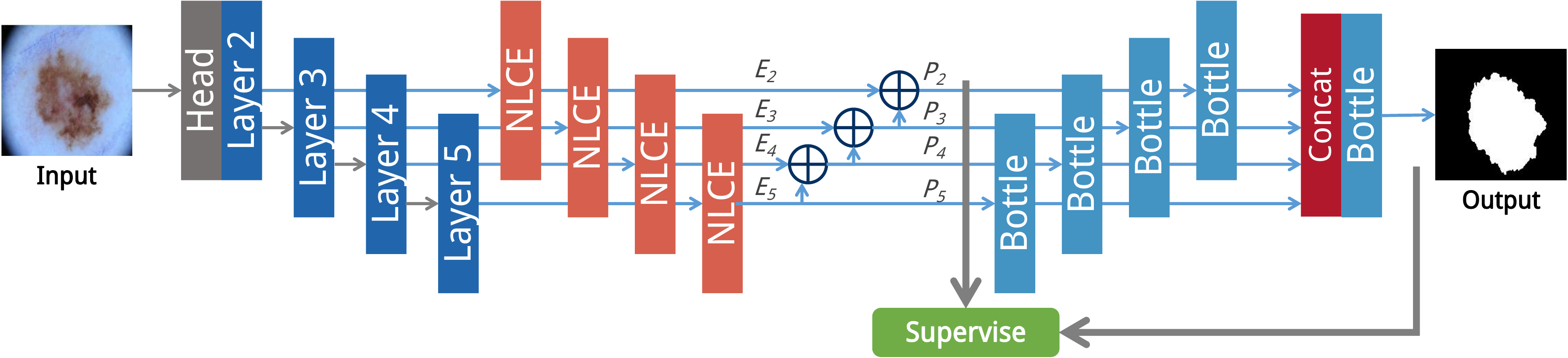}
  \end{center}
  \caption{The overall architecture of our proposed non-local context encoding network (NLCEN). The left part is based on a ResNet backbone and a feature pyramid. An NLCE module is added to bottom-up feature activations before lateral connections at different levels, and independent supervision is applied to predictions at all levels. The multi-scale information fused from all the pyramid features are used to refine the prediction and produce segmentation.\vspace{-4mm}}
  \label{fig:net}
\end{figure*}

\section{Methodology}
We propose a robust biomedical image segmentation module, called non-local context encoder (NLCE), against adversarial attacks. NLCEs capture short- and long-range spatial dependencies and strengthen the features with channel-wise feature map attention using the encoded global contexts. The effectiveness of global spatial dependencies and global contextual information contributes to the robustness of NLCE against attacks. In order to refine segmentation and capture sharp boundaries, we introduce coarse-to-fine non-local context encoding network (NLCEN), which captures the robust enhanced feature representations at different levels and then learns the fused multi-scale features. In this section, we introduce the NLCE module and the NLCEN framework in more detail.

\subsection{Non-Local Context Encoder}
Our non-local context encoder takes an $H \times W \times C$ feature map as input. It captures spatial short- and long-range dependencies in the feature map by following the design by \citeauthor{wang2018non} \citeyear{wang2018non}. It considers the feature map as a set of $C$-dimensional features $X = \{\mathbf{x}_1,...\mathbf{x}_N\}$, where $N= H\times W$ is the total number of features. We define the pairwise function $f$ that learns a relationship between any two features $\mathbf{x}_i$ and $\mathbf{x}_j$ as
\begin{equation}
f(\mathbf{x}_i, \mathbf{x}_j) = \exp\left(\theta(\mathbf{x}_i)^{T}\phi(\mathbf{x}_j)\right),
\end{equation}
where $\theta(\mathbf{x}_i) = W_{\theta}\mathbf{x}_i$ and $\phi(\mathbf{x}_j) = W_{\phi}\mathbf{x}_j$ are feature embeddings, where $W_{\theta}$ and $W_{\phi}$ are learned weight matrices.

The non-local response $\mathbf{y}_{i}$ for feature $\mathbf{x}_{i}$ is defined as
\begin{equation}
\mathbf{y}_{i} = \frac{1}{C(\mathbf{x})}\sum_{j=1}^{N}f(\mathbf{x}_{i}, \mathbf{x}_{j})g(\mathbf{x}_{j}),
\end{equation}
where the unary function $g$ is a mapping with a learned weight matrix $W_{g}$ to compute the representation $g(\mathbf{x}_j) = W_g\mathbf{x}_j$ of $\mathbf{x}_j$. $C(\mathbf{x})$ is the normalization factor, defined as $C(\mathbf{x}) = \sum_{i=1}^{N}f(\mathbf{x}_i,\mathbf{x}_j)$. The non-local response $\mathbf{y}_i$ captures short- and long-range dependencies via considering all features in the above non-local operation.

Next, the enhanced features $\mathbf{z}_i = W_z\mathbf{y}_i + \mathbf{x}_i$ ($W_z$ maps $\mathbf{y}_i$ to the $C$-dimensional space), which combine the non-local response $\mathbf{y}_i$ with the original feature $\mathbf{x}_i$, are fed into the context encoder discussed below. The feature map with the size of $H \times W \times C$ constructed from the enhanced features is denoted as $F_z$. Inspired by \citeauthor{zhang2018context} \citeyear{zhang2018context}, we learn a global codebook $D = \{\mathbf{d}_1,...\mathbf{d}_K\}$, which contains $K$ $C''$-dimensional codewords. The codebook represents global statistical information about the non-local enhanced features, and each codeword represents a visual center. We transform the enhanced features to the same dimensionality as the codewords via a $1 \times 1$ convolution, and the resulting $C''$-dimensional features are denoted as $Z' = \{\mathbf{z}'_1,...\mathbf{z}'_N \}$. The normalized residual $\mathbf{e}_{ik} $ between an enhanced feature $\mathbf{z}'_i$ and a codeword $\mathbf{d}_k$ is defined as
\begin{equation}
\mathbf{e}_{ik} = \frac{\exp\left( -s_k\|\mathbf{r}_{ik}\|^2 \right)}{R(\mathbf{e}_i)}\mathbf{r}_{ik},
\end{equation}
where $\mathbf{r}_{ik} = \mathbf{z}'_i - \mathbf{d}_k$ is the residual between feature $\mathbf{z}'_i$ and codeword $\mathbf{d}_k$, $s_k$ is a learned smoothing factor for codeword $\mathbf{d}_k$, and ${R(\mathbf{e}_i)} = \sum_{l=1}^{K}\exp(-s_l\|\mathbf{r}_{il}\|^2)$ is the normalization factor for feature $\mathbf{x}_i$. Thus, the residual information for all features captured by the codeword $\mathbf{d}_k$ is defined as $\mathbf{e}_k = \sum_{i=1}^{N}\mathbf{e}_{ik}$, and the global context is defined as $\mathbf{e} = \sum_{k=1}^{K}\sigma(\mathbf{e}_k)$, where $\sigma$ denotes Batch Normalization with ReLU.

Then, the global context $\mathbf{e}$ encoded from the spatial non-local features is used to strengthen the features using channel-wise feature map attention by predicting a channel-wise scaling factor $\gamma = \mbox{sigmoid}(W_\gamma\mathbf{e})$, where $W_\gamma$ is a learned weight matrix. The output from the NLCE module, $F_z \otimes \gamma$, is a channel-wise multiplication between the non-local enhanced feature map $F_z$ and the channel-wise scaling factor $\gamma$.

The architecture of our NLCE module is shown in Figure \ref{fig:nlce}. The NLCE module first captures short- and long-range spatial dependencies to denoise and strengthen the feature map, and then scales the feature map channels by scaling factors predicted using the encoded global context. Global dependencies and global contexts reduce the negative impact of adversarial noise, and give rise to the robustness of the NLCE module against adversarial attacks. Fusing information from the highly related pixels or the global context in forward propagation gradually weakens adversarial noise to a pixel or a semantic proposal.

\subsection{Non-Local Context Encoding Network (NLCEN)}
Our proposed coarse-to-fine non-local context encoding network (NLCEN) takes one biomedical image as input and produces a segmentation of organs or lesions at the pixel level. NLCEN has two phases, and its overall architecture is shown in Figure~\ref{fig:net}.

The architecture of the global phase is based on the ResNet backbone~\cite{he2016deep} and feature pyramid network. The fused information of low-level and high-level features by upsampling high-level features can capture rich contextual information with high resolution. An NLCE module is attached to the last residual block of conv2 through conv5 respectively to obtain multi-level robust non-local feature maps, denoted as $E_2,...,E_5$. Following FPN, the fused feature map $P_i$ ($i=2,3,4$) is obtained by element-wise addition between $E_i$ and the $1\times 1$ convolved and up-sampled $P_{i+1}$, and $P_5$ is obtained by attaching a $1\times 1$ convolutional layer to $E_5$. Feature maps $P_2,...,P_5$ are used to independently produce segmentation results by feeding each of them through a distinct $3\times 3$ convolution filter and a bilinear interpolation layer. Supervision is directly applied to each of these segmentation results.

Multi-level feature maps are fused together via upsampling and concatenation after going through bottleneck operations \cite{he2016deep}, and the refined segmentation prediction is produced directly from the fused feature map with multi-scale information. The number of bottleneck operations is respectively $0, 1, 2, 3$ for feature maps $P_2, P_3, P_4, P_5$. During testing, the final output is produced from the refined segmentation prediction. 

The loss function for a single map prediction is defined as the sum of cross-entropy losses at individual pixels between the ground truth and the predicted segmentation map:
\begin{equation}
L_s = \sum_i^{|I|} \log p_{i,g_i},
\end{equation}
where $|I|$ denotes the total number of pixels, $g_i$ is the ground-truth label at pixel $i$, $p_{i,g_i}$ is the probability that pixel $i$ is classified to category $g_i$.

We denote the loss for the segmentation predictions obtained from $P_2,...,P_5$ as $L_g^{2},...,L_g^{5}$, and the loss for the refined segmentation as $L_r$. The total loss is defined as:
\begin{equation}
L = \frac{1}{4}\sum_{i=2}^{5}L_g^{i} + \lambda L_r,
\end{equation}
where $\lambda=0.25$ is a weight balancing multiple coarse predictions from the global phase and the refined prediction from the refinement phase.

\section{Experimental Results}
\subsection{Datasets}
We have conducted evaluations on two commonly used benchmark biomedical image datasets, the Japanese Society of Radiological Technology (JSRT) dataset for lung segmentation~\cite{shiraishi2000development} and the International Symposium on Biomedical Imaging (ISBI 2016) dataset for skin lesion segmentation~\cite{gutman2016skin}.

The JSRT dataset was first introduced to help diagnostic training and testing for tuberculosis. It contains 154 nodule and 93 non-nodule post-anterior~(PA) chest radiographs with a $2048 \times 2048$ high resolution and wide density range. We split chest radiographs into a training set of 124 images and a test set of 123 images by following previous practices in the literature~\cite{hwang2017accurate}. The ground truth for the JSTR dataset is provided in \cite{van2006segmentation}.

The ISBI 2016 dataset provides 900 training images and 379 testing images with binary masks of skin lesion. The size of the images ranges from $524 \times 718$ to $2848 \times 4288$.

\subsection{Adversarial Attacks}
We adopt the target Iterative FGSM attack method~\cite{kurakin2016adversarial} to generate adversarial samples for a concrete model because the iterative white-box attacking methods have a high success rate. An attack sets the target as the inverse of ground-truth masks, denoted as $S_t$, and the adversarial sample of a single example in each iteration is defined as:
\begin{equation}
\mathbf{x}_{t+1}^{adv} = \mbox{clip}(\mathbf{x}_t^{adv} - \alpha \cdot \mbox{sign} (\nabla_{\mathbf{x}_t^{adv}}L_r(f(\mathbf{x}_{t}^{adv};\theta_f)), \epsilon),
\end{equation}
where $\mathbf{x}_0^{adv}$ is initialized to $\mathbf{x}$, the intensity of the adversarial perturbation is $\epsilon$, the step size of iterations is denoted as $\alpha$, and $\theta_f$ represents network parameters.

Following \citeauthor{kurakin2016adversarial}, we set $\alpha = 1$ , the number of iterations to $\min(\epsilon + 4, \lceil 1.25 \epsilon \rceil)$, and the $L_\infty$ norm of adversarial perturbation to intensity. We generate adversarial samples by setting adversarial intensity to every value from $\{0.5, 1, 2, 4, 6, 8, 10, 12, 14, 16, 18, 20, 22, 24, 26, 28, 30, 32\}$.

\subsection{Evaluation Metrics}
We evaluate the robustness of biomedical image segmentation methods by measuring the drop in segmentation accuracy after adding adversarial perturbations with different intensities to the original testing images. Dice's coefficient ($DIC$) and Jaccard similarity coefficient ($JSC$) are commonly used accuracy metrics in biomedical image segmentation. $DIC$ and $JSC$ are computed as follows:
\begin{equation}
DIC = \frac{2 \cdot TP}{2 \cdot TP + FN + FP},
\end{equation}
\begin{equation}
JSC = \frac{TP}{TP + FN + FP},
\end{equation}
where $TP$, $TN$, $FP$, $FN$ are the number of pixel-level true positives, true negatives, false positives, and false negatives, respectively.

\subsection{Implementation}
Our proposed NLCEN with NLCE modules has been implemented on the open source deep learning framework, PyTorch\cite{paszke2017automatic}. We follow the same experimental setups as in \citeauthor{hwang2017accurate} and \citeauthor{sarker2018slsdeep}. Horizontal flips, vertical flips and random rotations with $\pm10$ degrees are used as data augmentation operations on the ISBI 2016 dataset while no data augmentation is applied to the JSRT dataset during training. We set the mini-batch size to 8, and all input images are resized to $256 \times 256$. The Adam optimizer is adopted to update network parameters with the learning rate set to $0.001$ initially and reduced by $10\%$ whenever the training loss stops decreasing until $0.0001$. We use a weight decay of $0.0001$ and an exponential decay rate for the first moment estimates and the second moment estimates of $0.9$ and $0.999$ respectively. It takes $2$ hours to train a model on the JSRT dataset in a single NVIDIA TITAN GPU and $2$ more hours to generate adversarial samples for testing when an intensity of adversarial perturbation is given. The training and testing times on the ISBI 2016 dataset are $4$ hours respectively.

\subsection{Comparison with the State of the Art}
We compare the robustness of our proposed NLCEN with that of five state-of-the-art methods for lung segmentation and skin lesion segmentation, including dilated residual and pyramid pooling networks (SLSDeep)~\cite{sarker2018slsdeep}, network-wise training of convolutional networks (NWCN)~\cite{hwang2017accurate}, convolutional networks for biomedical image segmentation (UNet)~\cite{ronneberger2015u}, fully convolutional architectures for multi-class segmentation (InvertNet)~\cite{novikov2018fully} and segmentation with fully convolutional-deconvolutional networks (CDNN)~\cite{yuan2017automatic}. All the evaluations of the above networks are conducted on both the JSRT and ISBI 2016 datasets. According to the scale of the datasets, we adopt a ResNet-18 backbone for the JSRT dataset and a ResNet-50 backbone for the ISBI 2016 dataset. On each dataset, we first train a benchmark segmentation model on the training set and compute segmentation accuracy metrics ($DIC$ and $JSC$) on the testing set; and then, under each given intensity of perturbation, we generate adversarial samples of the testing set on the basis of the benchmark model and test its segmentation accuracy on the generated adversarial samples.

\begin{figure}[h]
  \includegraphics[width=\linewidth]{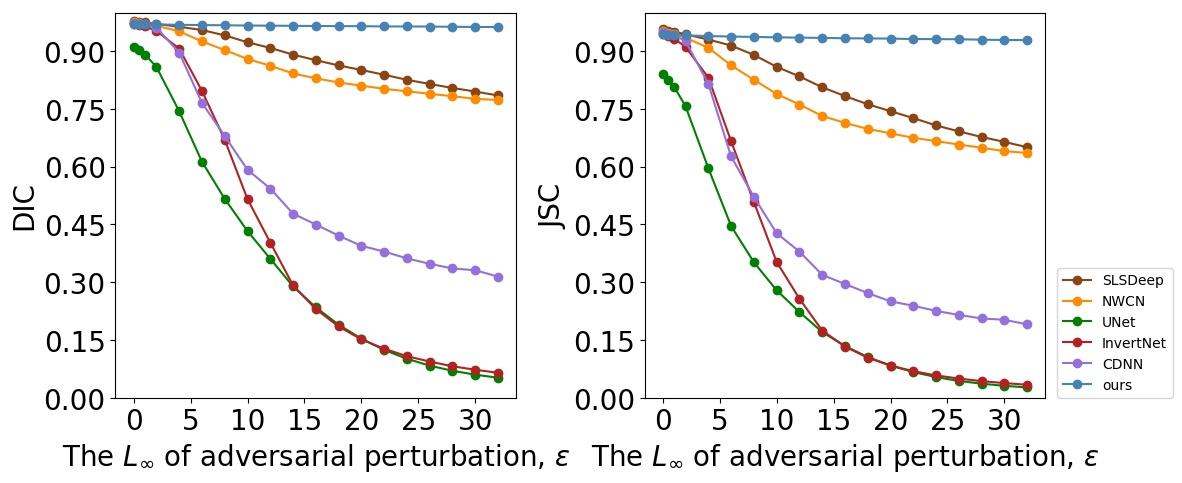}
  \caption{Comparison of quantitative results in terms of $DIC$ and $JSC$ on the JSRT lung segmentation dataset.}
  \label{fig:lungC}
\end{figure}

\begin{figure}[h]
  \includegraphics[width=\linewidth]{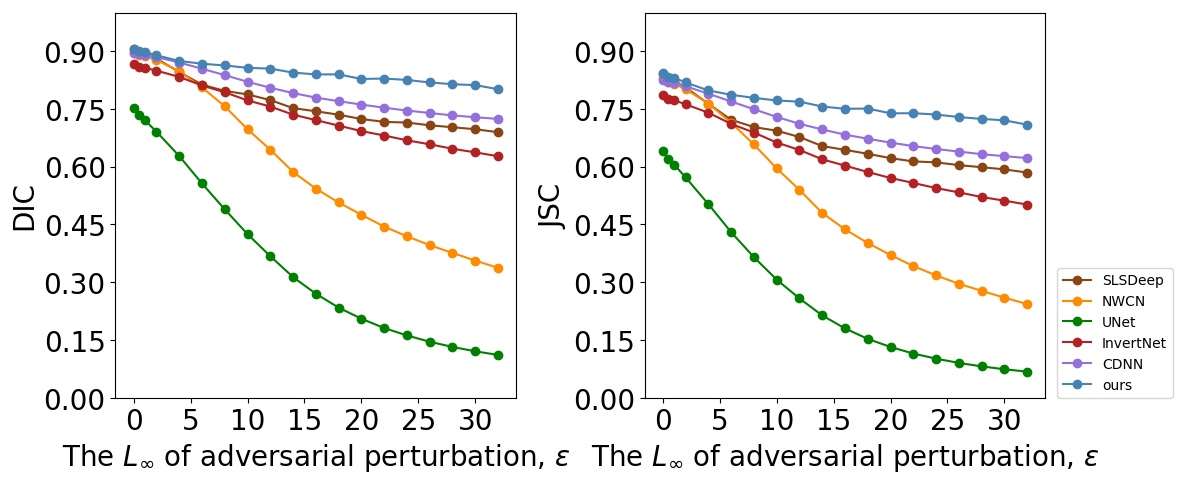}
  \caption{Comparison of quantitative results in terms of $DIC$ and $JSC$ on ISBI 2016 skin lesion segmentation dataset.\vspace{-4mm}}
  \label{fig:skinC}
\end{figure}

\textbf{Quantitative Evaluation} Figures~\ref{fig:lungC} and \ref{fig:skinC} show evaluation results in terms of $DIC$ and $JSC$ on the JSRT and ISBI 2016 datasets respectively. In these figures, we can find that NLCEN achieves the highest accuracy on clean skin lesion images, and achieves almost the top performance on clean lung images. Even when the strongest adversarial perturbation ($\epsilon = 32$) is exerted, it still maintains the highest accuracy. Its accuracy drops by only $0.01$ ($0.971$ to $0.963$) in $DIC$ and $0.02$ ($0.945$ to $0.929$) in $JSC$ on the JSRT dataset, and drops by $0.11$ ($0.907$ to $0.801$) in $DIC$ and $0.14$ ($0.844$ to $0.704$) in $JSC$ on the ISBI 2016 dataset. The results show that adversarial attacks have almost no effects on our lung segmentation model. The drop in accuracy on the ISBI 2016 skin dataset is larger than that on the JSRT dataset because there is very little contextual information in skin lesion images. Even though, our NLCEN is still the most robust one of all the models. Moreover, this experiment also indicates that the outstanding robustness of our model against adversarial samples with different levels of perturbation intensity.

\begin{figure}[h]
  \centering
  \includegraphics[width=0.9\linewidth]{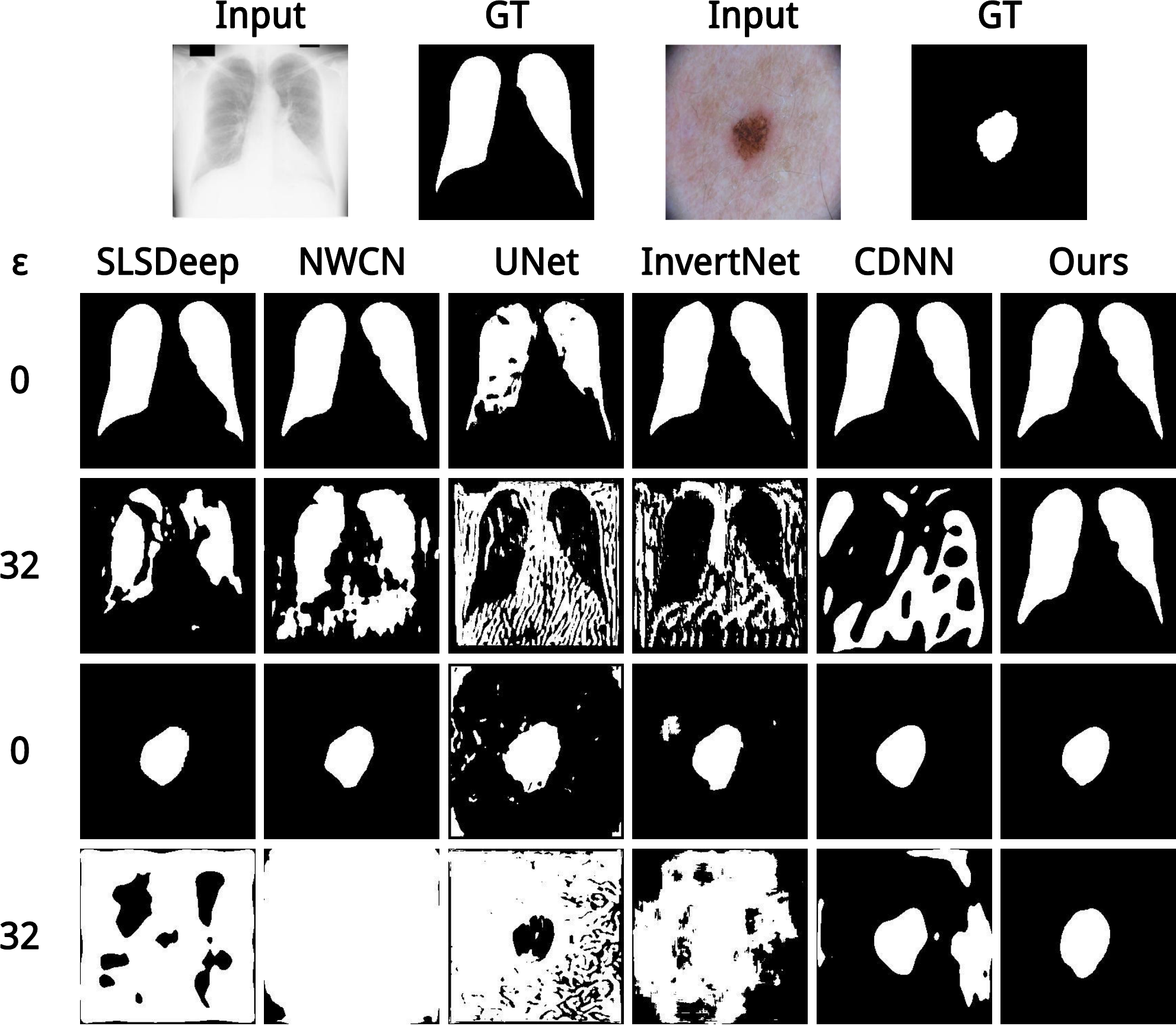}
  \caption{Comparison of segmentation results obtained from SLSDeep, NWCN, UNet, InvertNet, CDNN and our NLCEN when they are attacked by targeted Iterative FGSM with $\epsilon = 32$.\vspace{-4mm}}
  \label{fig:figQ}
\end{figure}

\textbf{Qualitative Evaluation} Figure~\ref{fig:figQ} visually compares segmentation results from our model and five existing methods when they are under the attack of targeted Iterative FGSM with $\epsilon = 32$. Refer to the supplementary materials for more results.

Our method achieves the most accurate segmentations among all the methods when they face adversarial attacks. Segmentation results from our method on adversarial samples are almost the same as those of our method on clean images, which demonstrates the robustness of our model against adversarial attacks. Under adversarial attacks, our method produces accurate segmentations of lung images, and segments out skin lesions completely; on the other hand, lung segmentations obtained from NWCN, UNet, InvertNet and CDNN are appalling, and all the other methods fail on skin lesion segmentation.

\subsection{Ablation Studies}
As discussed in the Methodology section, the robustness of our NLCE modules against adversarial attacks comes from global spatial dependencies and global contextual information. To verify their validity and necessity, we compare NLCEN with its three variants (i.e. NLCEN without NLCE modules (w/o NLCE), NLCEN without modeling global dependencies (w/o NL) and NLCEN without capturing global contexts (w/o CE)), which are trained and tested on the JSTR dataset. For the fairness of the comparison, we train the w/o NLCE model first. Then, we fine-tune the w/o NL, w/o CE and NLCEN models separately by freezing the layers of the w/o NLCE model. Finally, we fine-tune NLCEN without freezing any layer to obtain the fine-tuned model.

\begin{figure}[!b]
  \includegraphics[width=\linewidth]{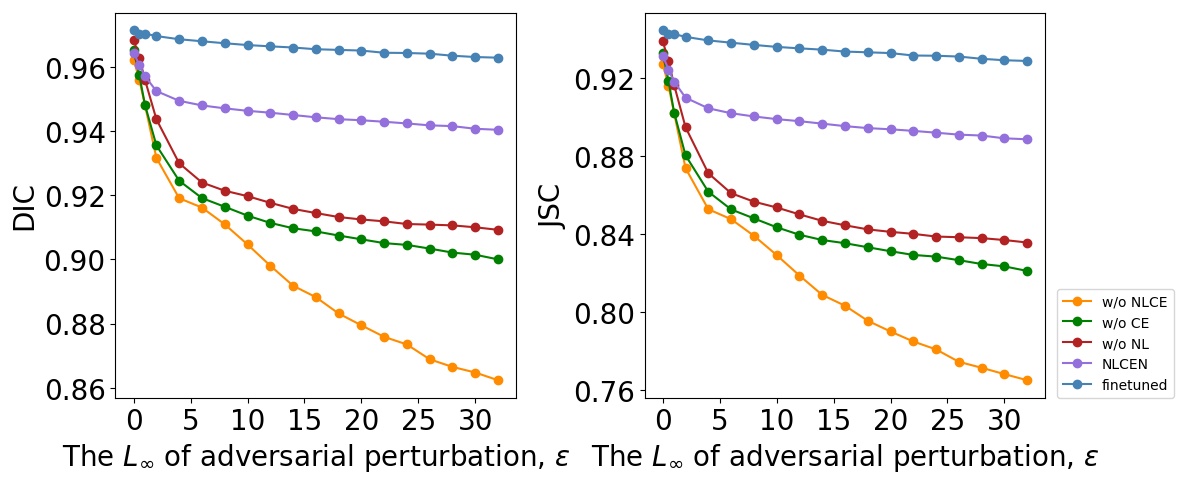}
  \caption{Ablation study on our non-local context encoding network.}
  \label{fig:LungA}
\end{figure}

The robustness of these models are evaluated and the results are shown in Figure~\ref{fig:LungA}. The non-local dependencies part or the global context part alone can already improve robustness, and the complete NLCE module with both parts can enhance the robustness further. That demonstrates the necessity of global dependencies and global contexts as well as the possibility of cooperation between them. In addition, the accuracy and robustness can be further enhanced by fine-tuning the NLCEN without freezing any layer.

\begin{figure}[h]
  \includegraphics[width=\linewidth]{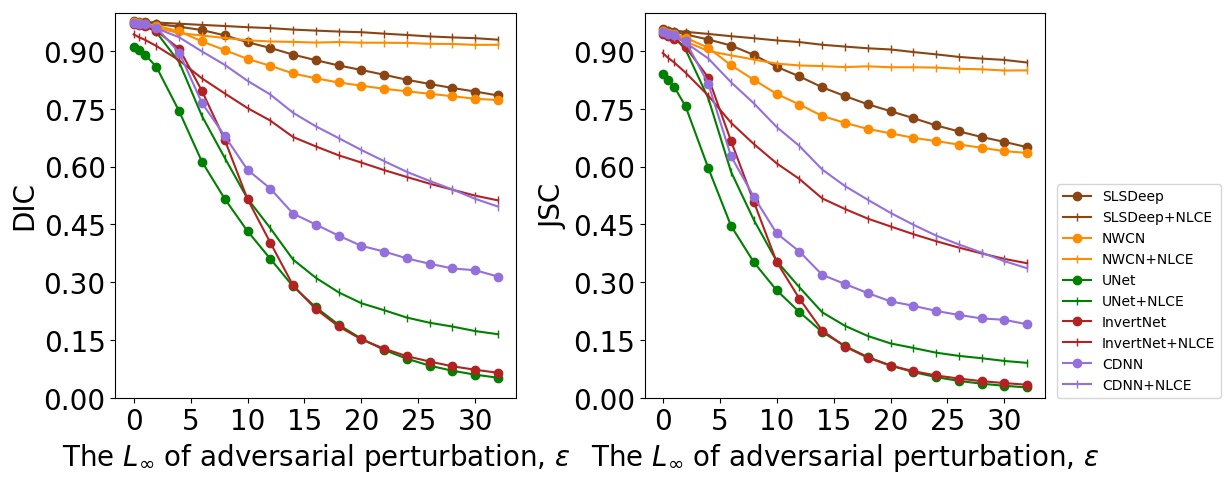}
  \caption{Comparison of robustness with and without non-local context encoder in other biomedical image segmentation methods.}
  \label{fig:LungU}
\end{figure}

\subsection{Generalization}
To verify that our non-local context encoder can be easily integrated into other networks, we instantiate an NLCE version for SLSDeep, NWCN, UNet, InvertNet and CDNN networks, respectively. Except for NWCN, we add one NLCE module between the last downsampling layer and the first upsampling layer in each network. For NWCN, we add two NLCE modules because it has two subnetworks. Finally, we train these updated networks from scratch and test those networks with NLCE modules on the JSRT dataset.

Figure~\ref{fig:LungU} shows a comparison between methods with NLCE modules and those without on the JSRT dataset. Methods with NLCE modules achieve significantly higher $DIC$ and $JSC$ than those without. This reveals NLCE modules are compatible with other biomedical image segmentation methods to strengthen their defense against adversarial attacks.

\section{Conclusions}
In this paper, we have proposed a non-local context encoder which is a robust biomedical image segmentation module against adversarial attacks. It is designed to not only capture global spatial dependencies by learning the response at a single feature as a weighted sum of all the features, but also strengthen the features with channel-wise feature map attention by using encoded global contextual information. The NLCE modules are core components of our non-local context encoding network (NLCEN) for robust and accurate biomedical image segmentation. 
Experimental results on both lung segmentation and skin lesion segmentation datasets have demonstrated that our proposed method can denoise adversarial perturbations and defend against adversarial attacks effectively while achieving accurate segmentation. In addition, our NLCE modules can help improve the robustness of other biomedical image segmentation methods against adversarial attacks.

\bibliography{aaai}

\begin{thebibliography}{}

\bibitem[\protect\citeauthoryear{Arnab, Miksik, and
  Torr}{2018}]{arnab2017robustness}
Arnab, A.; Miksik, O.; and Torr, P.~H.
\newblock 2018.
\newblock On the robustness of semantic segmentation models to adversarial
  attacks.
\newblock In {\em Proceedings of CVPR}.

\bibitem[\protect\citeauthoryear{Fukushima}{1980}]{fukushima1980neocognitron:}
Fukushima, K.
\newblock 1980.
\newblock Neocognitron: A self-organizing neural network model for a mechanism
  of pattern recognition unaffected by shift in position.
\newblock {\em Biological Cybernetics} 36(4):193--202.

\bibitem[\protect\citeauthoryear{Goodfellow, Shlens, and
  Szegedy}{2015}]{goodfellow2015explaining}
Goodfellow, I.; Shlens, J.; and Szegedy, C.
\newblock 2015.
\newblock Explaining and harnessing adversarial examples.
\newblock In {\em Proceedings of ICLR}.

\bibitem[\protect\citeauthoryear{Gutman \bgroup et al\mbox.\egroup
  }{2016}]{gutman2016skin}
Gutman, D.; Codella, N. C.~F.; Celebi, M.~E.; Helba, B.; Marchetti, M.~A.;
  Mishra, N.~K.; and Halpern, A.
\newblock 2016.
\newblock Skin lesion analysis toward melanoma detection: A challenge at the
  international symposium on biomedical imaging (isbi) 2016, hosted by the
  international skin imaging collaboration (isic).
\newblock {\em CoRR} abs/1605.01397.

\bibitem[\protect\citeauthoryear{He \bgroup et al\mbox.\egroup
  }{2016}]{he2016deep}
He, K.; Zhang, X.; Ren, S.; and Sun, J.
\newblock 2016.
\newblock Deep residual learning for image recognition.
\newblock In {\em Proceedings of CVPR},  770--778.

\bibitem[\protect\citeauthoryear{Hwang and Park}{2017}]{hwang2017accurate}
Hwang, S., and Park, S.
\newblock 2017.
\newblock Accurate lung segmentation via network-wise training of convolutional
  networks.
\newblock In {\em Deep Learning in Medical Image Analysis and Multimodal
  Learning for Clinical Decision Support}. Springer.
\newblock  92--99.

\bibitem[\protect\citeauthoryear{Kurakin, Goodfellow, and
  Bengio}{2016}]{kurakin2016adversarial}
Kurakin, A.; Goodfellow, I.; and Bengio, S.
\newblock 2016.
\newblock Adversarial machine learning at scale.
\newblock In {\em Proceedings of ICLR}.

\bibitem[\protect\citeauthoryear{LeCun, Bengio, and
  Hinton}{2015}]{lecun2015deep}
LeCun, Y.; Bengio, Y.; and Hinton, G.
\newblock 2015.
\newblock Deep learning.
\newblock {\em nature} 521(7553):436.

\bibitem[\protect\citeauthoryear{Lecun \bgroup et al\mbox.\egroup
  }{1989}]{lecun1989backpropagation}
Lecun, Y.; Boser, B.~E.; Denker, J.~S.; Henderson, D.; Howard, R.~E.; Hubbard,
  W.; and Jackel, L.~D.
\newblock 1989.
\newblock Backpropagation applied to handwritten zip code recognition.
\newblock {\em Neural Computation} 1(4):541--551.

\bibitem[\protect\citeauthoryear{Li and Yu}{2016}]{li2016visual}
Li, G., and Yu, Y.
\newblock 2016.
\newblock Visual saliency detection based on multiscale deep cnn features.
\newblock {\em IEEE Transactions on Image Processing} 25(11):5012--5024.

\bibitem[\protect\citeauthoryear{Li and Yu}{2018}]{li2018contrast}
Li, G., and Yu, Y.
\newblock 2018.
\newblock Contrast-oriented deep neural networks for salient object detection.
\newblock {\em IEEE Transactions on Neural Networks and Learning Systems}.

\bibitem[\protect\citeauthoryear{Li \bgroup et al\mbox.\egroup
  }{2016}]{li2016lstm}
Li, Z.; Gan, Y.; Liang, X.; Yu, Y.; Cheng, H.; and Lin, L.
\newblock 2016.
\newblock Lstm-cf: Unifying context modeling and fusion with lstms for rgb-d
  scene labeling.
\newblock In {\em Proceedings of ECCV},  541--557.
\newblock Springer.

\bibitem[\protect\citeauthoryear{Li \bgroup et al\mbox.\egroup
  }{2017}]{li2017instance}
Li, G.; Xie, Y.; Lin, L.; and Yu, Y.
\newblock 2017.
\newblock Instance-level salient object segmentation.
\newblock In {\em Proceedings of CVPR},  247--256.

\bibitem[\protect\citeauthoryear{Li \bgroup et al\mbox.\egroup
  }{2018}]{li2018cross}
Li, G.; Gan, Y.; Wu, H.; Xiao, N.; and Lin, L.
\newblock 2018.
\newblock Cross-modal attentional context learning for rgb-d object detection.
\newblock {\em IEEE Transactions on Image Processing}.

\bibitem[\protect\citeauthoryear{Lin \bgroup et al\mbox.\egroup
  }{2017}]{lin2017feature}
Lin, T.-Y.; Doll{\'a}r, P.; Girshick, R.~B.; He, K.; Hariharan, B.; and
  Belongie, S.~J.
\newblock 2017.
\newblock Feature pyramid networks for object detection.
\newblock In {\em Proceedings of CVPR}, volume~1, ~4.

\bibitem[\protect\citeauthoryear{Litjens \bgroup et al\mbox.\egroup
  }{2017}]{litjens2017survey}
Litjens, G.; Kooi, T.; Bejnordi, B.~E.; Setio, A. A.~A.; Ciompi, F.;
  Ghafoorian, M.; van~der Laak, J.~A.; Van~Ginneken, B.; and S{\'a}nchez, C.~I.
\newblock 2017.
\newblock A survey on deep learning in medical image analysis.
\newblock {\em Medical image analysis} 42:60--88.

\bibitem[\protect\citeauthoryear{Long, Shelhamer, and
  Darrell}{2015}]{long2015fully}
Long, J.; Shelhamer, E.; and Darrell, T.
\newblock 2015.
\newblock Fully convolutional networks for semantic segmentation.
\newblock In {\em Proceedings of CVPR},  3431--3440.

\bibitem[\protect\citeauthoryear{Novikov \bgroup et al\mbox.\egroup
  }{2018}]{novikov2018fully}
Novikov, A.~A.; Lenis, D.; Major, D.; Hladuvka, J.; Wimmer, M.; and B{\"u}hler,
  K.
\newblock 2018.
\newblock Fully convolutional architectures for multi-class segmentation in
  chest radiographs.
\newblock {\em IEEE Transactions on Medical Imaging}.

\bibitem[\protect\citeauthoryear{Paszke \bgroup et al\mbox.\egroup
  }{2017}]{paszke2017automatic}
Paszke, A.; Gross, S.; Chintala, S.; Chanan, G.; Yang, E.; DeVito, Z.; Lin, Z.;
  Desmaison, A.; Antiga, L.; and Lerer, A.
\newblock 2017.
\newblock Automatic differentiation in pytorch.
\newblock In {\em NIPS-W}.

\bibitem[\protect\citeauthoryear{Ren \bgroup et al\mbox.\egroup
  }{2015}]{ren2015faster}
Ren, S.; He, K.; Girshick, R.; and Sun, J.
\newblock 2015.
\newblock Faster r-cnn: Towards real-time object detection with region proposal
  networks.
\newblock In {\em Proceedings of NIPS},  91--99.

\bibitem[\protect\citeauthoryear{Ronneberger, Fischer, and
  Brox}{2015}]{ronneberger2015u}
Ronneberger, O.; Fischer, P.; and Brox, T.
\newblock 2015.
\newblock U-net: Convolutional networks for biomedical image segmentation.
\newblock In {\em Proceedings of International Conference on Medical image
  computing and computer-assisted intervention},  234--241.
\newblock Springer.

\bibitem[\protect\citeauthoryear{Sarker \bgroup et al\mbox.\egroup
  }{2018}]{sarker2018slsdeep}
Sarker, M.; Kamal, M.; Rashwan, H.~A.; Banu, S.~F.; Saleh, A.; Singh, V.~K.;
  Chowdhury, F.~U.; Abdulwahab, S.; Romani, S.; Radeva, P.; et~al.
\newblock 2018.
\newblock S{LSD}eep: Skin lesion segmentation based on dilated residual and
  pyramid pooling networks.
\newblock In {\em Proceedings of International Conference on Medical Image
  Computing and Computer Assisted Intervention}.

\bibitem[\protect\citeauthoryear{Shiraishi \bgroup et al\mbox.\egroup
  }{2000}]{shiraishi2000development}
Shiraishi, J.; Katsuragawa, S.; Ikezoe, J.; Matsumoto, T.; Kobayashi, T.;
  Komatsu, K.-i.; Matsui, M.; Fujita, H.; Kodera, Y.; and Doi, K.
\newblock 2000.
\newblock Development of a digital image database for chest radiographs with
  and without a lung nodule: receiver operating characteristic analysis of
  radiologists' detection of pulmonary nodules.
\newblock {\em American Journal of Roentgenology} 174(1):71--74.

\bibitem[\protect\citeauthoryear{Szegedy \bgroup et al\mbox.\egroup
  }{2013}]{szegedy2013intriguing}
Szegedy, C.; Zaremba, W.; Sutskever, I.; Bruna, J.; Erhan, D.; Goodfellow, I.;
  and Fergus, R.
\newblock 2013.
\newblock Intriguing properties of neural networks.
\newblock {\em arXiv preprint arXiv:1312.6199}.

\bibitem[\protect\citeauthoryear{Tram{\`e}r \bgroup et al\mbox.\egroup
  }{2018}]{tramer2017ensemble}
Tram{\`e}r, F.; Kurakin, A.; Papernot, N.; Goodfellow, I.; Boneh, D.; and
  McDaniel, P.
\newblock 2018.
\newblock Ensemble adversarial training: Attacks and defenses.
\newblock In {\em Proceedings of ICLR}.

\bibitem[\protect\citeauthoryear{Van~Ginneken, Stegmann, and
  Loog}{2006}]{van2006segmentation}
Van~Ginneken, B.; Stegmann, M.~B.; and Loog, M.
\newblock 2006.
\newblock Segmentation of anatomical structures in chest radiographs using
  supervised methods: a comparative study on a public database.
\newblock {\em Medical image analysis} 10(1):19--40.

\bibitem[\protect\citeauthoryear{Wang \bgroup et al\mbox.\egroup
  }{2018}]{wang2018non}
Wang, X.; Girshick, R.; Gupta, A.; and He, K.
\newblock 2018.
\newblock Non-local neural networks.
\newblock In {\em Proceedings of CVPR}.

\bibitem[\protect\citeauthoryear{Xie \bgroup et al\mbox.\egroup
  }{2017}]{xie2017adversarial}
Xie, C.; Wang, J.; Zhang, Z.; Zhou, Y.; Xie, L.; and Yuille, A.
\newblock 2017.
\newblock Adversarial examples for semantic segmentation and object detection.
\newblock In {\em Proceedings of ICCV}.

\bibitem[\protect\citeauthoryear{Yuan}{2017}]{yuan2017automatic}
Yuan, Y.
\newblock 2017.
\newblock Automatic skin lesion segmentation with fully
  convolutional-deconvolutional networks.
\newblock {\em arXiv preprint arXiv:1703.05165}.

\bibitem[\protect\citeauthoryear{Zhang \bgroup et al\mbox.\egroup
  }{2018}]{zhang2018context}
Zhang, H.; Dana, K.; Shi, J.; Zhang, Z.; Wang, X.; Tyagi, A.; and Agrawal, A.
\newblock 2018.
\newblock Context encoding for semantic segmentation.
\newblock In {\em Proceedings of CVPR}.

\end{thebibliography}
\bibliographystyle{aaai}
\end{document}